\documentclass[10pt,twocolumn,letterpaper]{article}

\usepackage{cvpr}              

\usepackage{xcolor}
\usepackage{pifont} 
\newcommand{\cmark}{\textcolor{green!60!black}{\ding{51}}}
\newcommand{\xmark}{\textcolor{red!70!black}{\ding{55}}}
\definecolor{cvprblue}{rgb}{0.21,0.49,0.74}
\usepackage[pagebackref,breaklinks,colorlinks,allcolors=cvprblue]{hyperref}

\def\paperID{} 
\def\confName{CVPR}
\def\confYear{2026}

\title{%
  \textbf{StereoWorld: Geometry-Aware Monocular-to-Stereo Video Generation}
}
\author{Ke Xing\textsuperscript{1 2} ,\quad Xiaojie Jin\textsuperscript{1 $\dagger$ $\ddagger$} ,\quad Longfei Li\textsuperscript{1} ,\quad Yuyang Yin\textsuperscript{1} ,\quad Hanwen Liang\textsuperscript{3} ,\quad Guixun Luo\textsuperscript{1} \\ Chen Fang\textsuperscript{2} ,\quad Jue Wang\textsuperscript{2} ,\quad
Konstantinos N. Plataniotis\textsuperscript{3}  ,\quad Yao Zhao\textsuperscript{1} ,\quad Yunchao Wei\textsuperscript{1 $\dagger$}\\
\normalsize
\textsuperscript{1}Beijing Jiaotong University ,\quad
\textsuperscript{2}Dzine AI ,\quad
\textsuperscript{3}University of Toronto \\
\normalsize
$\dagger$ Corresponding Author ,\quad $\ddagger$ Project Lead \\
\normalsize
\href{https://ke-xing.github.io/StereoWorld/}{\textbf{https://ke-xing.github.io/StereoWorld/}}
\vspace{-1em}
}

\begin{document}

\twocolumn[{
\renewcommand\twocolumn[1][]{#1}
\maketitle
\begin{center}
    \includegraphics[width=1.0\textwidth]{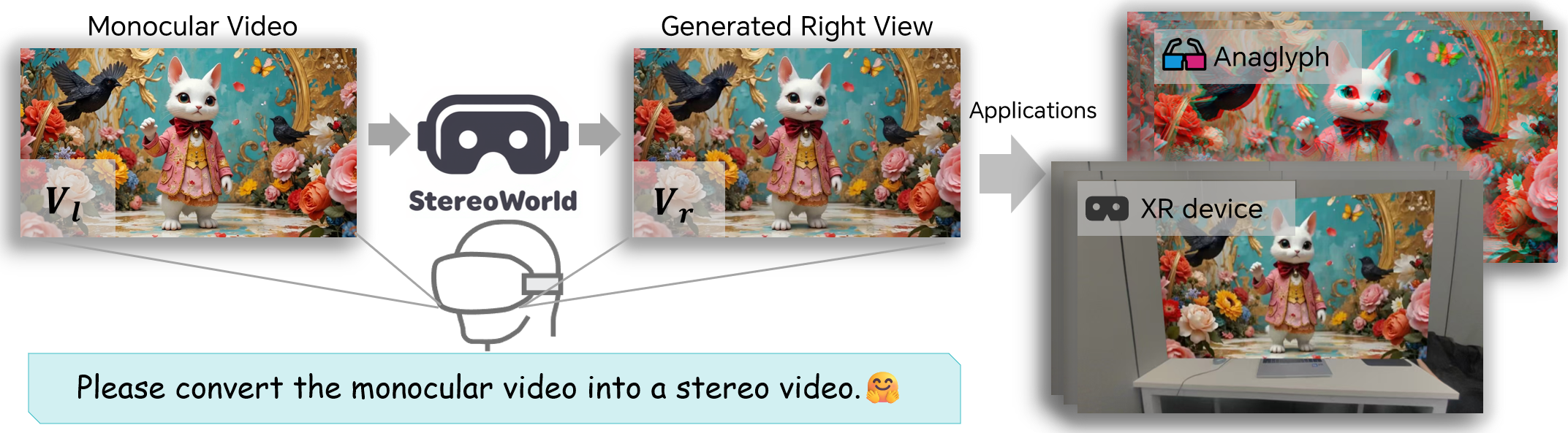}
    \captionof{figure}{\textbf{Examples generated by StereoWorld.} Our method directly generates stereo videos from arbitrary monocular videos without requiring additional information. The results can be displayed on 3D glasses, XR headsets, and other stereoscopic devices. }

    \label{fig:teaser}
\end{center}
}]

\begin{abstract}
The growing adoption of XR devices has fueled strong demand for high-quality stereo video, yet its production remains costly and artifact-prone.
To address this challenge, we present \textbf{StereoWorld, an end-to-end framework} that repurposes a pretrained video generator for high-fidelity monocular-to-stereo video generation. Our framework jointly conditions the model on the monocular video input while explicitly supervising the generation with a \textbf{geometry-aware regularization} to ensure 3D structural fidelity.
A spatio-temporal tiling scheme is further integrated to enable efficient, high-resolution synthesis.
To enable large-scale training and evaluation, we curate a high-definition stereo video dataset containing over 11M frames aligned to natural human interpupillary distance (IPD).
Extensive experiments demonstrate that StereoWorld substantially outperforms prior methods, generating stereo videos with superior visual fidelity and geometric consistency.


\end{abstract}
\section{Introduction}
\label{sec:intro}

The rapid adoption of Extended Reality (XR) devices, such as the Apple Vision Pro and Meta Quest, has fueled growing demand for immersive stereo video. However, producing high-quality stereo footage relies on dedicated dual-camera rigs with tight calibration and synchronization, making it inaccessible to most creators. Given the vast abundance of monocular videos online, a scalable algorithm that can transform ordinary monocular footage into realistic, high-fidelity stereo video would be highly beneficial.

Existing approaches for monocular-to-stereo conversion can be broadly grouped into two paradigms.
The first treats the task as novel-view synthesis (NVS) via 3D scene reconstruction. Traditional SfM pipelines \cite{schonberger2016structure} and modern neural rendering methods such as NeRF \cite{mildenhall2021nerf} and 3D Gaussian Splatting (3DGS) \cite{kerbl20233d} attempt to recover scene geometry and camera parameters before rendering a new perspective for the right eye. However, this pipeline is vulnerable to pose inaccuracies and unconstrained scene dynamics.
Alternative pose-free models~\cite{wang2024dust3r,zhang2024monst3r,wang2025vggt} also struggle with the geometric ambiguities and non-rigid motions in real-world videos.
Therefore, these approaches often generate stereos with unstable geometry and temporal inconsistency.

An alternative and more recent paradigm is the depth-warping-inpainting pipeline, powered by advances in diffusion models~\cite{dai2024svg, shi2024stereocrafter, wang2024stereodiffusion, zhao2024stereocrafter, zhang2024spatialme}.
Depth is first estimated from the input video and used to warp frames into the target viewpoint; occluded regions are then hallucinated by an inpainting model to plausibly fill in the missing areas.
While conceptually simple, this pipeline suffers from distinct drawbacks. Specifically, the inpainting phase is decoupled from stereo geometry estimation, breaking pixel-level correspondence and resulting in texture distortions, color shifts, and stereo artifacts that degrade viewing comfort.

To address these limitations, we introduce \textbf{StereoWorld}, a novel end-to-end diffusion-based framework that converts a general monocular video generative model into a high-fidelity stereo generator.
Instead of relying on fragile pose estimation or multi-stage warping pipelines, we leverages the rich spatio-temporal priors of the foundational video model to explicitly learn stereo geometry and generate coherent right-eye views directly from monocular input.

Our framework achieves this with the following proposed techniques. 
First, we extend the base video diffusion model with \textbf{monocular-conditioning} that allows the model to incorporate a monocular video as strong guidance. 
Second, to overcome the geometric inaccuracies of prior methods, we introduce a novel \textbf{geometry-aware regularization} strategy composed of a disparity and depth supervision. 
Disparity supervision enforces accurate stereo correspondence, mitigating cross-view misalignment and temporal disparity drift to improve stability and visual comfort.
To further supplement geometric information, the model jointly diffuses RGB videos and their associated depth maps, explicitly learning 3D structure and providing stronger geometric guidance than RGB reconstruction alone.
Furthermore, a \textbf{spatio-temporal tiling strategy} enables the efficient generation of high-resolution, long-duration videos. This optimization allows our framework to overcome the typical constraints of diffusion models, making it scalable for producing practical, high-fidelity content suitable for modern XR displays.

Another major challenge for stereo generation is the lack of suitable training data.
Existing datasets feature \textit{baselines~\footnote{In stereo vision, the \textit{baseline} is the precise physical distance between the optical centers of the two camera lenses.}} that far exceed the human interpupillary distance (IPD). This wide \textit{baseline} is unsuitable for XR devices as it leads to exaggerated parallax, which can easily cause visual discomfort for the viewer.
To overcome this limitation, we curate a large-scale, high-resolution stereo video benchmark dataset aligned to human IPD, as summarized in Tab.~\ref{tab:dataset_stats}, enabling both reliable training and fair evaluation.

\begin{table}[tbp]
\centering
\small
\setlength{\tabcolsep}{2pt} 
\begin{tabular}{lcccc}
\toprule
Dataset &  Domain & IPD-aligned & Available & Frames \\
\midrule
Spring~\cite{mehl2023spring}          & Optical Flow & \xmark & \cmark  & 5K \\
Sintel~\cite{butler2012naturalistic}          & Optical Flow & \xmark & \cmark  & 1K \\
VKITTI2~\cite{cabon2020virtual}              & Driving & \xmark & \cmark  & 21K \\
PLT-D3~\cite{tokarsky2024plt}              & Driving & \xmark & \cmark  & 3K \\
IRS~\cite{wang2019irs}           & Robotics  & \xmark & \cmark  & 103K \\
TartanAir~\cite{wang2020tartanair}        & Robotics  & \xmark & \cmark  & 306K \\
3D Movies~\cite{ranftl2020towards}      & Moives & \cmark & \xmark  & 75K \\
\midrule
StereoWorld-11M   & Moives & \cmark & \cmark  & \textbf{11M} \\
\bottomrule
\end{tabular}
\caption{\textbf{Comparison of the stereo datasets.} Existing datasets are generally not IPD-aligned (e.g., Spring, VKITTI2), while datasets that are IPD-aligned are not publicly available (e.g., 3D Movies). Our StereoWorld is the first large-scale, IPD-aligned dataset.}
\vspace{-1.3em}
\label{tab:dataset_stats}
\end{table}

Our main contributions are summarized as follows:
\begin{itemize}
    \item 
    We propose \textbf{StereoWorld}, the first fully end-to-end diffusion framework that adapts a pretrained monocular video generative model into a stereo generator with high visual fidelity and geometric accuracy.
    
    \item 
    We build a large-scale, high-definition \textbf{stereo video dataset} aligned with human-IPD, featuring over 11M curated Blu-ray SBS video frames across diverse genres with comprehensive evaluation metrics.
    \item 
    Extensive experiments demonstrate that StereoWorld substantially outperforms prior works in visual quality, geometric consistency, and temporal stability, with clear advantages in objective metrics and subjective perception.

\end{itemize}

\section{Related Work}
\label{sec:related work}

\subsection{Novel View Synthesis}
With the advent of deep learning, novel view synthesis (NVS) has progressed rapidly, evolving from traditional Structure-from-Motion (SfM) \cite{schonberger2016structure} methods to neural approaches such as NeRF \cite{mildenhall2021nerf} and 3D Gaussian Splatting (3DGS) \cite{kerbl20233d}. 
While NeRF implicitly models scene geometry, 3DGS introduces explicit representations that improve reconstruction quality and efficiency.
Recent works extend NVS to dynamic scenes and feed-forward reconstruction.
For example, 4DGS \cite{wu20244d} and Shape of Motion \cite{wang2024shape} reconstruct dynamic geometry from monocular videos, while methods like MVSplat \cite{chen2024mvsplat}, PixelSplat \cite{charatan2024pixelsplat}, and VGGT \cite{wang2025vggt} leverage large pretrained models for fast, pose-free reconstruction. 
However, reconstructed geometry remains sparse, and the visual fidelity of novel views is still limited, restricting their applicability to stereo video generation.

\subsection{Diffusion-based Stereo Generation}
Existing diffusion-based stereo generation methods—whether training-free approaches such as SVG \cite{dai2024svg}, StereoCrafter-Zero \cite{shi2024stereocrafter}, T-SVG \cite{jin2024t}, and StereoDiffusion \cite{wang2024stereodiffusion}, or pretrained models including StereoCrafter \cite{zhao2024stereocrafter}, SpatialMe \cite{zhang2024spatialme}, StereoConversion \cite{mehl2024stereo}, SpatialDreamer \cite{lv2025spatialdreamer}, ImmersePro \cite{shi2024immersepro}, ReStereo \cite{huang2025restereo}, and GenStereo \cite{qiao2025genstereo}—generally follow a similar pipeline: monocular depth estimation, view warping, and diffusion-based inpainting of occluded regions.
But this paradigm disrupts the natural video distribution, causing spatial–temporal inconsistencies and degraded fidelity.
In contrast, our method departs fundamentally from this paradigm by directly generating stereo videos in an end-to-end manner, thereby ensuring cross-view consistency and preserving visual fidelity.

\subsection{Video Diffusion Models}
Diffusion models~\cite{ho2020denoising,song2020denoising,sohl2015deep,song2020score}  have achieved remarkable success in image generation~\cite{rombach2022high,saharia2022photorealistic,nichol2021glide,podell2023sdxl,wu2025qwen,cao2025hunyuanimage}, inspiring their extension to 3D domain \cite{poole2022dreamfusion,xiang2025structured,zhao2025hunyuan3d,feng2025seed3d} and the video domain~\cite{ho2022video,blattmann2023stable,wan2025wan,hong2022cogvideo,yang2024cogvideox,brooks2024video}. Early works \cite{ho2022video} trained diffusion models directly on video datasets, while subsequent approaches augmented pretrained image diffusion models with temporal modules to capture motion dynamics \cite{blattmann2023stable,guo2023animatediff}. More recently, native video diffusion architectures built upon 3D-VAEs, such as Sora \cite{brooks2024video} and CogVideo \cite{hong2022cogvideo}, have demonstrated impressive spatio-temporal coherence and high-quality synthesis. The strong generative capacity and temporal consistency of pretrained video diffusion models make them a promising foundation for tasks like monocular-to-stereo video generation.

\begin{figure*}[htbp]
    \centering 
    \includegraphics[width=\textwidth]{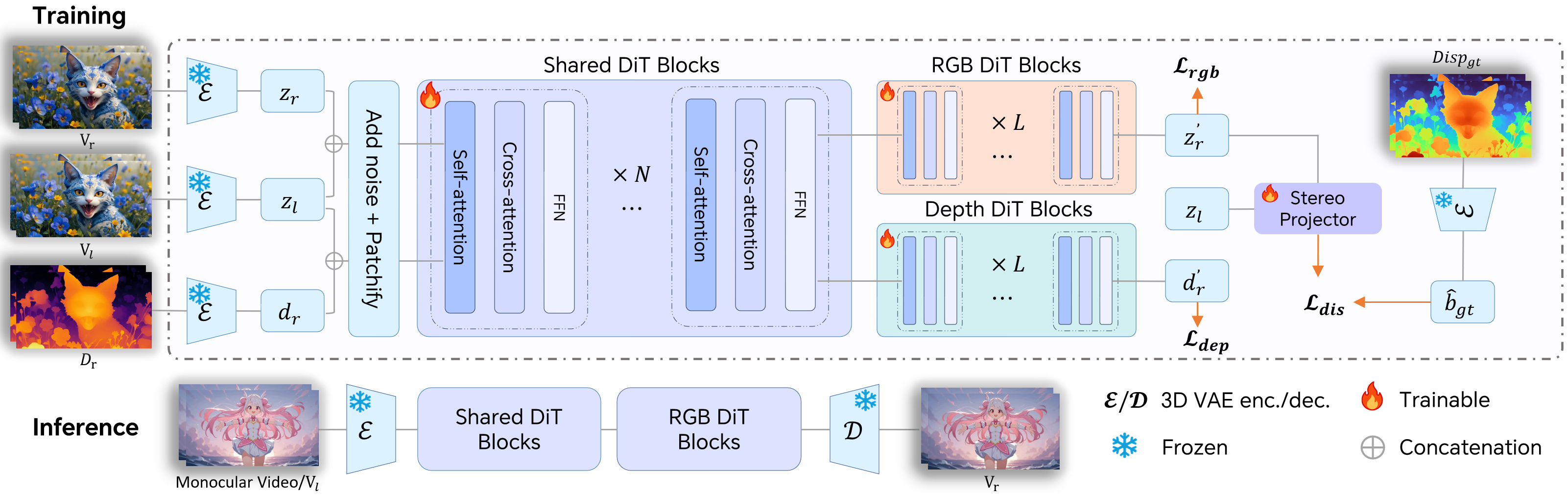} 

    \caption{\textbf{Overall framework of StereoWorld.} Before training, we use Video Depth Anything~\cite{chen2025video} and Stereo Any Video~\cite{jing2025stereo} to obtain the depth maps $D_r$ and disparity maps ${Disp}_\text{gt}$, and the left-view videos are then concatenated with the right-view videos and corresponding depth maps along the frame dimension in the latent space as conditioning inputs. During training, a lightweight differentiable stereo projector estimates the disparity between the input left-view and the generated right-view, which is supervised by disparity maps ${Disp}_\text{gt}$ via disparity loss to enforce accurate geometric correspondence. Additionally, the last few DiT blocks are duplicated to form dual branches, allowing the model to learn RGB and depth distributions separately to further supplement geometric information. During inference, only the shared and RGB DiT blocks are used, taking the monocular video as the sole input.}
    
    \vspace{-1.5em}
    \label{fig:framework} 
\end{figure*}

\section{Methodology}

In this section, we first review the fundamentals of video diffusion models (Sec.~\ref{ssec:preliminary}). We then detail our benchmark dataset construction (Sec.~\ref{ssec:dataset}), present the core training strategy (Sec.~\ref{ssec:training}), and describe practical optimizations for scalable generation (Sec.~\ref{ssec:optimization}).

\subsection{Preliminary: Video Diffusion Model}
\label{ssec:preliminary}
We build our framework on a pretrained text-to-video diffusion model based on the Diffusion Transformer(DiT) architecture~\cite{peebles2023scalable}. This model first uses a 3D Variational Auto-Encoder (3D VAE) to encode videos into a latent representation. Then DiT blocks integrate self-attention and cross-attention modules to jointly capture spatio–temporal dependencies and text–video interactions.
The model is trained under the Rectified Flow framework \cite{lipman2022flow}, where the forward process defines a linear trajectory between the data distribution and a standard normal distribution:
\begin{equation} 
    z_t = (1 - t)z_0 + t\epsilon, 
    \label{e1} 
\end{equation} 
where $\epsilon \sim \mathcal{N}(0, I)$ and $t$ denotes the diffusion timestep. 
The goal is to learn a velocity field $v_\Theta$ parameterized by the network weights $\Theta$, which transforms random noise $z_1 \sim p_1$ into data samples $z_0 \sim p_0$ through an ordinary differential equation (ODE): $\frac{dz_t}{dt} = v_\Theta(z_t, t) \label{e2} $. During training, we minimize the Conditional Flow Matching (CFM) loss \cite{esser2024scaling} by regressing the target vector field $u_t$ that generates a valid probability path between $p_0$ and $p_1$: 
\begin{equation} 
    \mathbb{E}_{t, p_t(z, \epsilon), p(\epsilon)} || v_\Theta(z_t, t) - u_t(z_0|\epsilon) ||_2^2, 
    \label{e3} 
\end{equation} 
where $u_t(z,\epsilon) := \psi_t'(\psi_t^{-1}(z|\epsilon)|\epsilon)$, and $\psi(\cdot|\epsilon)$ represents the mapping defined in Eq.~\ref{e1}.

\subsection{StereoWorld-11M Dataset Construction}
\label{ssec:dataset}

High-fidelity stereo video generation models are critically dependent on large-scale, high-quality training data. However, existing datasets are ill-suited for generating stereo video optimized for the human eye. These datasets \cite{butler2012naturalistic,cabon2020virtual,wang2019irs,geiger2012we,karaev2023dynamicstereo,wang2020tartanair,mehl2023spring,tokarsky2024plt} are primarily developed for applications like depth estimation, autonomous driving, or robotics. Therefore, their \textit{baseline} often exceeds 10 cm—far beyond the typical human IPD (55–75 mm), which can cause visual discomfort or dizziness when viewed stereoscopically.

To address this gap, we curated a new dataset tailored for stereo video generation with \textit{baseline} aligned to natural human perception.
We collected and cleaned over a hundred high-definition Blu-ray side-by-side (SBS) stereo movies from the Internet, spanning diverse genres such as animation, realism, war, sci-fi, historical, and drama, ensuring both visual diversity and richness for training.
All videos are unified into the SBS format by stretching and horizontally cropping to obtain left–right views, each with a resolution of 1080p, 16:9 aspect ratio, and 24 fps frame rate.
To match the training requirements of our base model (480p resolution, 81-frame inputs), we uniformly downscale each video to 480p.
To enhance motion diversity and increase temporal information density, we uniformly sample 81 frames per clip at fixed intervals.

\subsection{Framework}
\label{ssec:training}

Our primary objective is to generate a corresponding right-view video $V_r \in \mathbb{R}^{c \times f \times h \times w}$ from the left-view video $V_l \in \mathbb{R}^{c \times f \times h \times w}$.
Since existing video generative models are inherently monocular and lack this capability. To address this limitation, we adapt a pretrained monocular video generator to our stereo synthesis task through a simple yet effective conditioning way.

\noindent \textbf{Monocular-conditioning.}~~ The first challenge in adapting a foundational monocular video generator for stereo synthesis is devising an effective conditioning way. The model must generate a geometrically consistent right-view $V_r$ conditioned on the provided left-view $V_l$. The predominant existing paradigm is a multi-stage depth-warping-inpainting pipeline~\cite{zhao2024stereocrafter,huang2025restereo,dai2024svg}. This approach lacks effective reference and fusion of information from the original left view during the inpainting process, leading to degraded visual quality in the generated results. A more integrated alternative, such as injecting left-view features via cross-attention, avoids this issue but requires significant architectural modifications and adds substantial computational overhead.

Inspired by ReCamMaster~\cite{bai2025recammaster}, we employ a simple yet effective conditioning strategy that aggregates left and right-view latents along the frame dimension. This allows the diffusion model to leverage its existing attention mechanisms to fuse information across space, time, and viewpoint simultaneously. Specifically, we encode the left and right videos into latent space via the VAE encoder $\mathcal{E}$: $z_l = \mathcal{E}(V_l)$ and $z_r = \mathcal{E}(V_r)$, where $z_l, z_r \in \mathbb{R}^{c' \times f' \times h' \times w'}$. We then concatenate these latents along the frame dimension, $z_i = [z_l, z_r]_{\text{frame-dim}},$
where $z_i \in \mathbb{R}^{b \times c' \times 2f' \times h' \times w'}$ to serve as the direct input to the diffusion model , as illustrated in Fig. \ref{fig:framework}. This strategy is highly efficient, requiring no architectural changes , as the model's existing 3D spatio-temporal self-attention layers naturally fuse information by operating across all tokens from both views.


The essence of stereoscopy lies in the depth variations among scene objects, which define their hierarchical spatial relationships relative to the observer.
However, relying solely on monocular conditioning and a standard $\mathcal{L}_{\text{rgb}}$ reconstruction loss is insufficient to capture such geometric structure.
As shown in Fig.~\ref{fig:ablation}, we observe that the model struggles to implicitly learn complex geometric structures from only RGB data, resulting in outputs with weak stereo perception, such as flattened object boundaries or unstable disparities. This suggests that the model requires a more explicit signal to guide its understanding of 3D geometry. Therefore, we propose a geometry-aware regularization strategy to enhance the model's 3D perception capacity. 

\begin{figure}[tbp]
    \centering 
    \includegraphics[width=\linewidth]{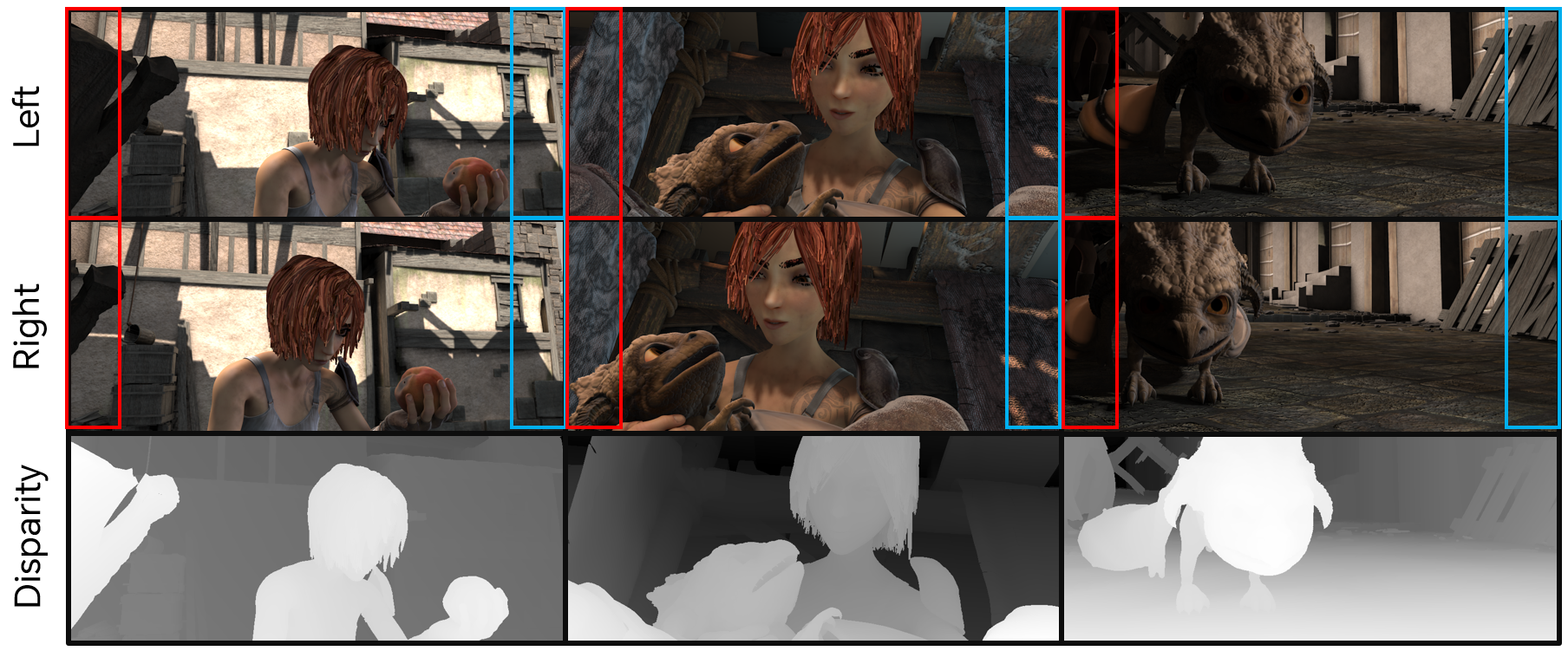} 

    \caption{\textbf{Non-overlapping regions between stereo views.} Horizontal camera translation introduces non-overlapping content, which disparity supervision alone cannot constrain, motivating the use of depth-based supervision. }

    \label{fig:struct} 
\end{figure}

\noindent \textbf{Geometry-aware Regularization.}
Our geometry-aware regularization consists of two complementary components—disparity and depth supervision—jointly designed to enhance stereo correspondence and geometric fidelity during right-view generation.

\noindent - Disparity supervision.
To enforce accurate stereo correspondence and mitigate cross-view misalignment and temporal disparity drift, we introduce a disparity-based loss.
First, we pre-compute the ground-truth disparity map $\hat{b}_\text{gt}$ by applying a pre-trained stereo matching network~\cite{jing2025stereo} to the ground-truth left ($V_l$) and right ($V_r$) video frames. This provides a geometrically accurate target.
During training, after the model predicts the denoised right-view latent $z_r'$, we employ a lightweight, differentiable stereo projector $\kappa$ to estimate the predicted disparity $\hat{b}_\text{pred}$ (Fig. \ref{fig:framework}). This projector takes the original left-view latent $z_l$ and the generated right-view latent $z_r'$ as input, $\hat{b}_\text{pred} = \kappa(z_l, z_r').$
The overall disparity loss is defined as
\begin{equation}
    \mathcal{L}_\text{dis} = \mathcal{L}_\text{log} + \lambda_\text{l1} \mathcal{L}_\text{l1},
\end{equation}
\newline
where $\lambda_\text{l1}$ is a weighting hyperparameter. The two loss terms are defined as: $ \mathcal{L}_\text{log} = \mathbb{E}\big[d^2\big] - \lambda_\text{1} \big(\mathbb{E}[d]\big)^2 $ that enforces global geometric consistency across disparity, $ \mathcal{L}_\text{l1} = \mathbb{E}[|\hat{b}_\text{pred} - \hat{b}_\text{gt}|] $ that penalizes pixel-wise disparity errors, \text{where } $ d = \log \hat{b}_\text{pred} - \log \hat{b}_\text{gt} $. Together, they explicitly guide the model to learn the stereo correspondence between left and right views, producing geometrically consistent stereo videos.

Although disparity maps encode geometric cues, they only capture correspondences within the overlapping regions between left and right views. As illustrated in Fig.~\ref{fig:struct}, when the camera translates horizontally to capture the right view, new regions appear on one side while others disappear on the opposite side. Since stereo matching operates only on overlapping content, the resulting disparity supervision provides incomplete geometric guidance for right-view generation. To address this limitation, we introduce an additional depth-based constraint.

\begin{figure}[tbp]
    \centering 
    \includegraphics[width=\linewidth]{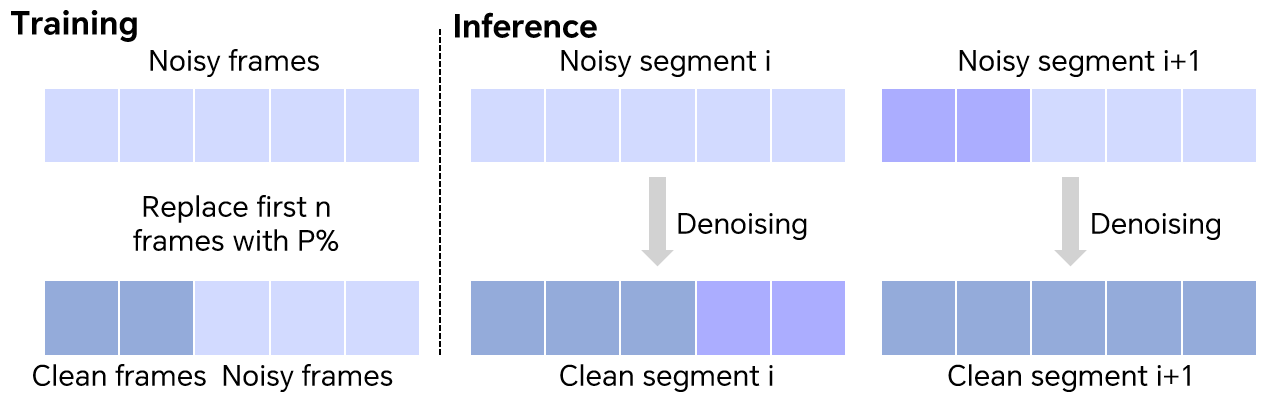} 

    \caption{\textbf{Temporal tiling strategy.} During training, the first few frames of noisy latents are replaced with ground-truth frames with a probability $p$. During inference, long videos are split into overlapping segments, with the last frames of the previous segment used to guide the next, ensuring temporal consistency.}

    \label{fig:long} 
\end{figure}

\noindent - Depth supervision.
To compensate for the missing geometric cues in non-overlapping regions, we introduce a depth-based supervision. Unlike disparity, depth provides a complete per-pixel geometric description of the target regions, including areas invisible to stereo matching.
We predict the right-view depth maps and constrain the model to generate them consistently with the RGB frames.
This reformulates the generation as a multi-objective joint prediction problem, where the model is simultaneously guided to learn the velocity field for both the RGB video and its corresponding depth map.
Specifically, we denote the latent representation of the right-view depth map as $d_r$. Following the same Conditional Flow Matching logic as Sec.~\ref{ssec:preliminary}, we define a new objective, $\mathcal{L}_\text{dep}$, to train the model to predict the velocity field for this depth distribution. We denote the updated model parameters as $\Theta'$. This results in a revised training objective with two main components:
\begin{align}
    \mathcal{L}_\text{rgb} &= \mathbb{E}_{t, p_t(z, \epsilon), p(\epsilon)}|| v_{\Theta'}(z_t, t) - u_t(z_0|\epsilon) ||_2^2, \\
    \mathcal{L}_\text{dep} &= \mathbb{E}_{t, p_t(d, \epsilon), p(\epsilon)}|| v_{\Theta'}(d_t, t) - u_t(d_0|\epsilon) ||_2^2.
\end{align}
To provide the target $d_r$ for this objective, we first pre-compute a per-frame depth map $D_r \in \mathbb{R}^{c \times f \times h \times w}$ for each right-view video $V_r$ using a state-of-the-art depth estimation model \cite{chen2025video}. This depth map is then encoded into its latent representation $d_r = \mathcal{E}(D_r)$ using the same VAE encoder.

A naive architectural approach for this multi-objective prediction would be to use the exact same set of DiT parameters (i.e., full parameter sharing) to learn the velocity fields for both RGB ($z_r$) and depth ($d_r$). However, this approach can hinder convergence and reduce learning efficiency, as the model is forced to reconcile potentially conflicting optimization gradients from two different data distributions within a single set of weights. To address this, we implement a specialized network architecture, as shown in Fig.~\ref{fig:framework}, that balances shared representation learning with task-specific refinement. We keep the initial transformer blocks shared, allowing the model to capture joint texture and geometric representations from both tasks. We then duplicate the weights of the final few DiT blocks to create two specialized branches: one dedicated to predicting the RGB velocity field and the other for the depth velocity field. This design enables the model to build a robust, structured understanding of both scene layout and depth hierarchy, leading to more geometrically accurate synthesis.

\begin{figure}[tbp]
    \centering 
    \includegraphics[width=\linewidth]{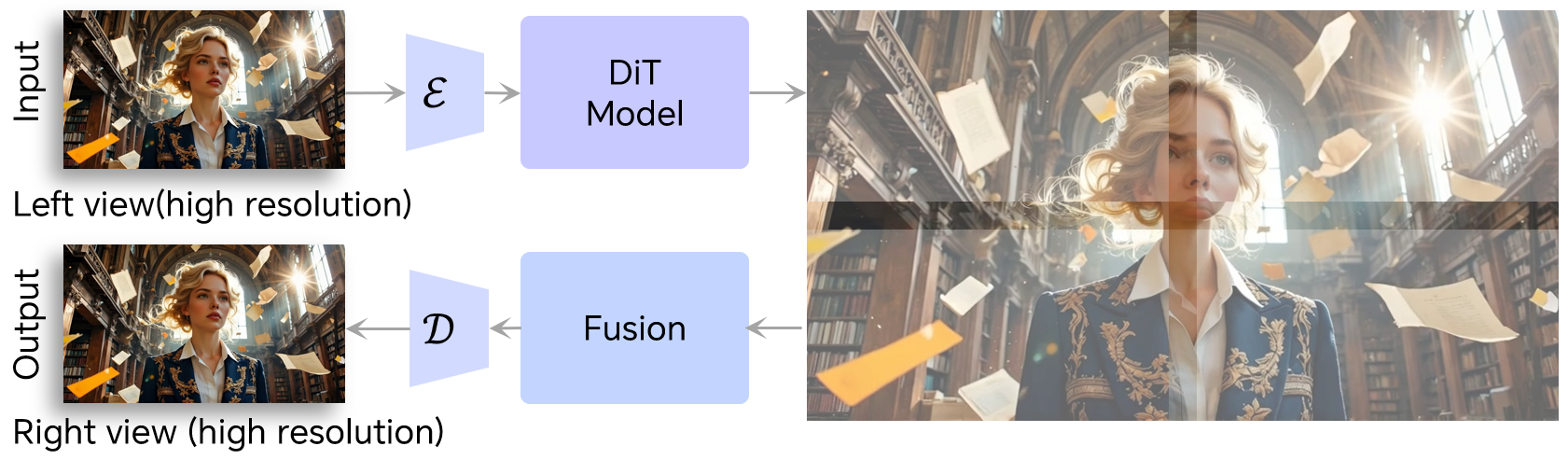} 

    \caption{\textbf{Spatial tiling strategy.} During inference, high-resolution videos are encoded into latents, which are split into overlapping tiles. Each tile is denoised independently, and then the tiles are stitched back to the original size with overlapping regions fused before decoding.}

    \label{fig:big} 
\end{figure}

\noindent \textbf{Training Objectives.} 
The overall training objective jointly supervises RGB reconstruction, depth consistency, and stereo disparity learning:
\begin{equation}
\mathcal{L} = \mathcal{L}_\text{rgb} + \mathcal{L}_\text{dep} + \lambda_\text{dis} \mathcal{L}_\text{dis},
\end{equation}
This integrated objective promotes both visual fidelity and geometric correctness, leading to more perceptually coherent and stereoscopically realistic video generation.

\begin{figure*}[tbp]
    \centering 
    \includegraphics[width=\textwidth]{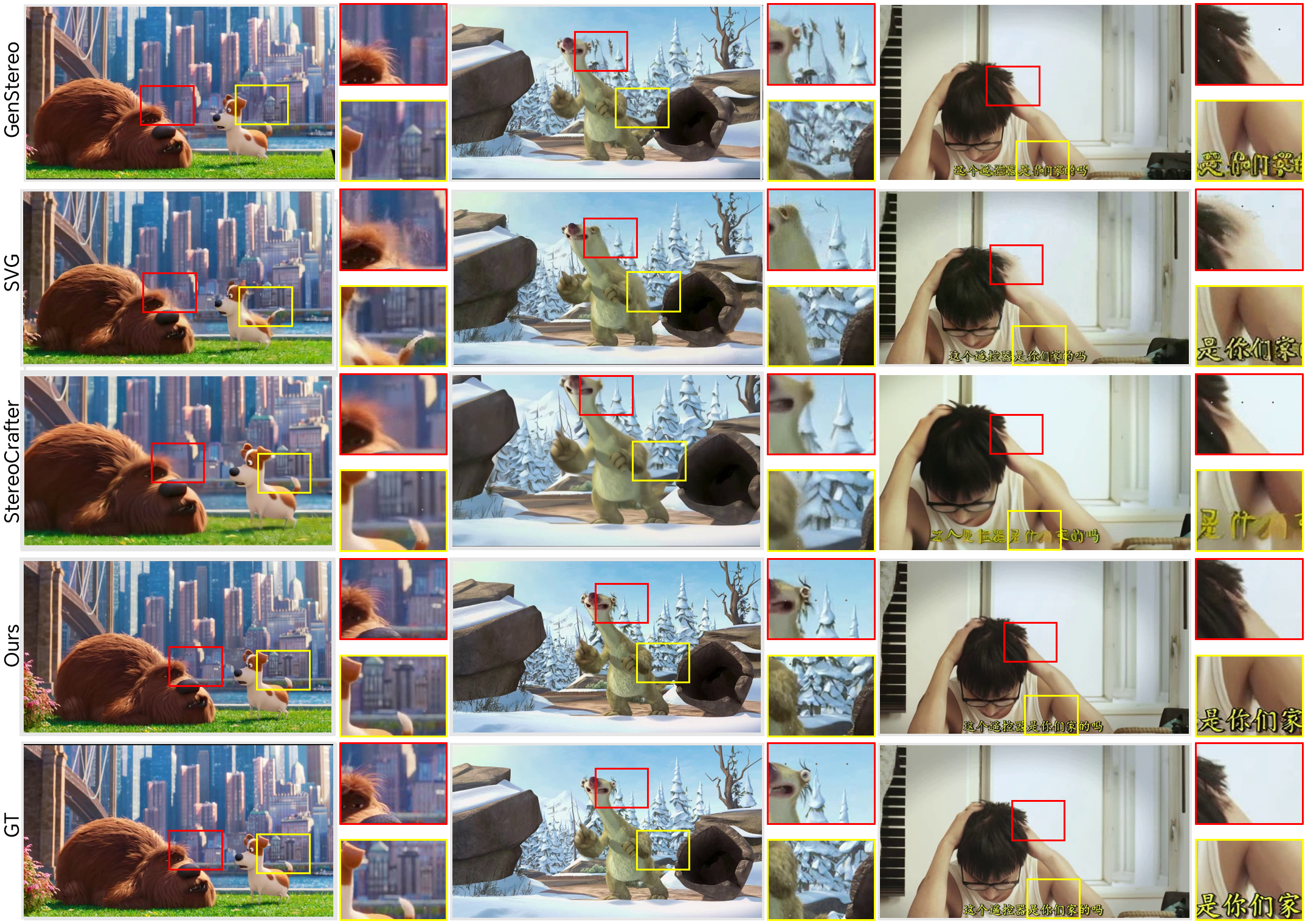}
    \caption{\textbf{Qualitative comparisons with state-of-the-art methods.} It shows that our method achieves the best generation quality, preserving fine details while maintaining strong visual consistency with the left view. Crucially, our method achieves far better text rendering quality than all baselines.} 
    \label{fig:comparison} 
\end{figure*}

\begin{table*}[t]
    \centering
    \begin{tabular*}{\textwidth}{@{\extracolsep{\fill}} l c c c c c c c} 
        \toprule
         Method &  PSNR ↑ &  SSIM ↑ &  LPIPS ↓ &  IQ-Score ↑&   TF-Score ↑ & EPE ↓ & D1-all ↓ \\
        \midrule
        GenStereo \cite{qiao2025genstereo} & 19.4486 & 0.6803 & 0.3008 & 0.4047 & 0.9642 & 35.0022 & 0.8954 \\
        SVG \cite{dai2024svg} & 18.0256 & 0.5881 & 0.3467 & 0.4714 & \textbf{0.9706} & 33.2508 & 0.9630 \\
        StereoCrafter \cite{zhao2024stereocrafter} & 23.0372 & 0.6561 & 0.1869 & 0.4370 & 0.9685 & 24.7784 & 0.5271 \\
        Ours & \textbf{25.9794} & \textbf{0.7964} & \textbf{0.0952} & \textbf{0.5019} & 0.9704 & \textbf{17.4527} & \textbf{0.4213}\\
        \bottomrule
    \end{tabular*}
    
    \caption{Quantitative comparisons with state-of-the-art methods on visual quality and geometry accuracy. IQ-Score and TF-Score refer to image quality and temporal flickering scores from VBench~\cite{huang2024vbench}. }
    \label{tab:comparison}

\end{table*}

\subsection{Practical and Scalable Optimization.}
\label{ssec:optimization}

\textbf{Temporal Tiling Strategy.}
Our base model generates only short clips (81 frames, about 3s at 24 FPS). To handle longer videos, we split them into overlapping segments, using the last frames of each segment to guide the next for smooth transitions \cite{zhao2024stereocrafter}. To further reduce flickering, during training we probabilistically replace first few frames of noisy latents with clean frames (probability $p$), enabling the model to learn robust long-range temporal consistency (Fig.~\ref{fig:long}).

\begin{figure*}[tbp]
    \centering 
    \includegraphics[width=\textwidth]{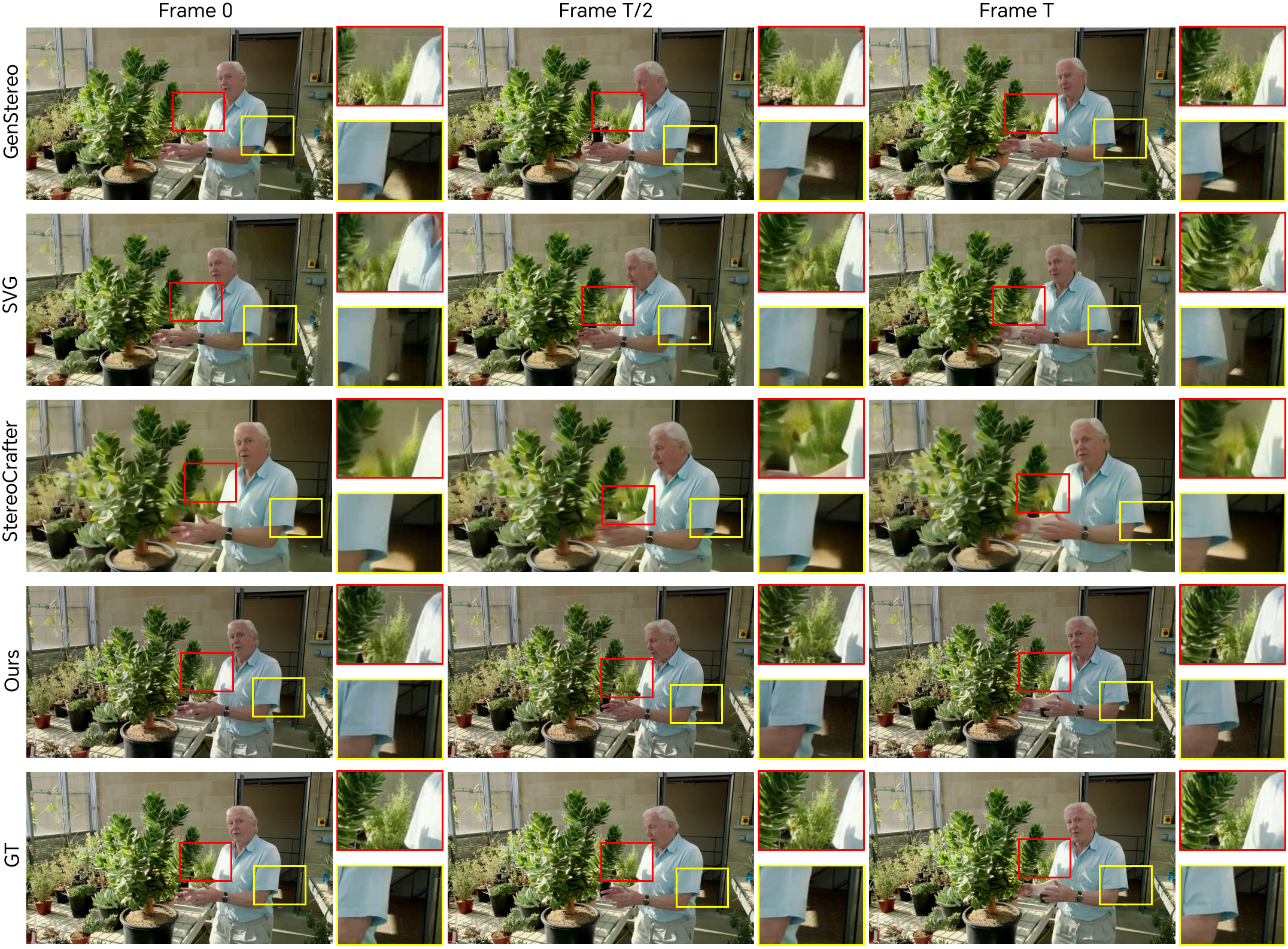}

    \caption{\textbf{Qualitative comparisons with state-of-the-art methods in the temporal dimension.} Our method maintains superior temporal consistency while preserving high visual quality and fine-grained detail fidelity compared to other methods.} 
    \label{fig:comparison2} 
\end{figure*}

\noindent \textbf{Spatial Tiling Strategy.}
Our model is trained at 480p. To handle high-resolution videos beyond the 480p training resolution, we adopt block-wise latent diffusion \cite{zhao2024stereocrafter}. High-resolution latents are split into overlapping tiles, each denoised independently, then stitched and fused before decoding (Fig.~\ref{fig:big}). This enables efficient high-resolution synthesis while preserving spatial details and visual coherence.

\section{Experiments}

\subsection{Experimental Setup}
\textbf{Implementation Details.}
We build our method upon Wan2.1-T2V-1.3B \cite{wan2025wan}, which generates 5-second video clips at 16 FPS with a spatial resolution of 832×480. 
To obtain depth supervision, we employ Video Depth Anything \cite{chen2025video} to estimate per-frame depth maps for all training videos. 
We further employ Stereo Any Video \cite{jing2025stereo} to generate ground-truth disparity maps used for supervising the disparity during training.
During fine-tuning, we adopt the LoRA \cite{hu2022lora} framework with a rank of 128, setting $\lambda_{\text{1}}=\lambda_{\text{l1}} = 0.1$ and $\lambda_{\text{dis}} = 0.5$.
The learning rate is set to $1 \times {10}^{-4}$, and the model is trained for one epoch, totaling approximately 9k optimization steps. 
Training is performed on 8 NVIDIA A800 GPUs using the AdamW optimizer \cite{loshchilov2017decoupled} under bfloat16 precision. The entire training process takes approximately 11 days to complete.

\noindent \textbf{Dataset.}
We train our model on the web-collected video dataset aforementioned. 
After preprocessing, the dataset contains 142,520 video clips, each with a spatial resolution of 480×832 and 81 frames, corresponding to approximately 7 seconds at 12 FPS. 
We randomly select 1,000 clips as the test set, with the remaining clips used for training.

\noindent \textbf{Evaluation Metrics.}
To quantitatively assess the generated right-view videos, we adopt PSNR~\cite{hore2010image}, SSIM~\cite{wang2004image}, and LPIPS~\cite{zhang2018unreasonable} to measure the generation fidelity with respect to the ground-truth right views. 
In addition, we evaluate image quality (IQ-Score) and temporal flickering (TF-Score) from VBench \cite{huang2024vbench} to measure visual quality and temporal consistency.
For disparity-level evaluation, we employ Stereo Any Video \cite{jing2025stereo} to estimate disparity maps from both the ground-truth stereo pairs and the generated pairs. 
We then compute EPE (End-Point-Error)~\cite{barron1994performance}—the average pixel-wise disparity error—and D1-all~\cite{geiger2012we}, which denotes the percentage of pixels whose disparity error exceeds a given threshold (typically 3 pixels or 5\% of the true disparity). 
These metrics together provide a comprehensive assessment of both visual fidelity and geometric accuracy.

\noindent \textbf{Baselines.}
Our approach targets video-to-video stereo generation, and several related methods have not released official implementations, we select three representative baselines for comparison. 
Specifically, we use GenStereo \cite{qiao2025genstereo} as a training-based image-to-image baseline, SVG \cite{dai2024svg} as a training-free video-to-video baseline, and StereoCrafter \cite{zhao2024stereocrafter} as a training-based video-to-video baseline.



\subsection{Comparisons}

\textbf{Qualitative Results.}
As shown in Fig.~\ref{fig:comparison} and Fig.~\ref{fig:comparison2}, the image-based method GenStereo fails to generalize to video data, exhibiting severe temporal instability and frame-wise distortions. The training-free method SVG struggles to inpaint occluded regions introduced by warping, producing visible artifacts and incomplete structures in the synthesized right views. While StereoCrafter generates visually coherent results, it tends to oversmooth fine textures, resulting in noticeable loss of high-frequency details. 
In contrast, our method achieves the most visually faithful and temporally stable results, accurately preserving scene geometry and fine-grained textures while maintaining strong semantic alignment with the left view. 
Most notably, StereoWorld excels in text rendering (a particularly challenging case for stereo generation) maintaining sharpness, legibility, and consistent spatial placement across both views, where all other baselines exhibit blurring or ghosting artifacts.

\noindent \textbf{Quantitative Results.}
As presented in Tab.~\ref{tab:comparison}, the image-based baseline GenStereo and the training-free method SVG obtain the lowest overall scores, consistent with the qualitative observations. Although StereoCrafter achieves competitive results on perceptual quality metrics, it exhibits substantially higher errors on geometry-related measures such as EPE and D1-all, indicating inaccurate disparity estimation and weaker stereo correspondence. 
In contrast, our method consistently outperforms all baselines across both visual and geometric metrics, achieving the best balance between visual quality and geometry accuracy. 
These results highlight that StereoWorld not only enhances the realism and temporal coherence of generated videos but also produces geometrically consistent stereo pairs that align more closely with human interpupillary distance.

\begin{figure}[tbp]
    \centering 
    \includegraphics[width=\linewidth]{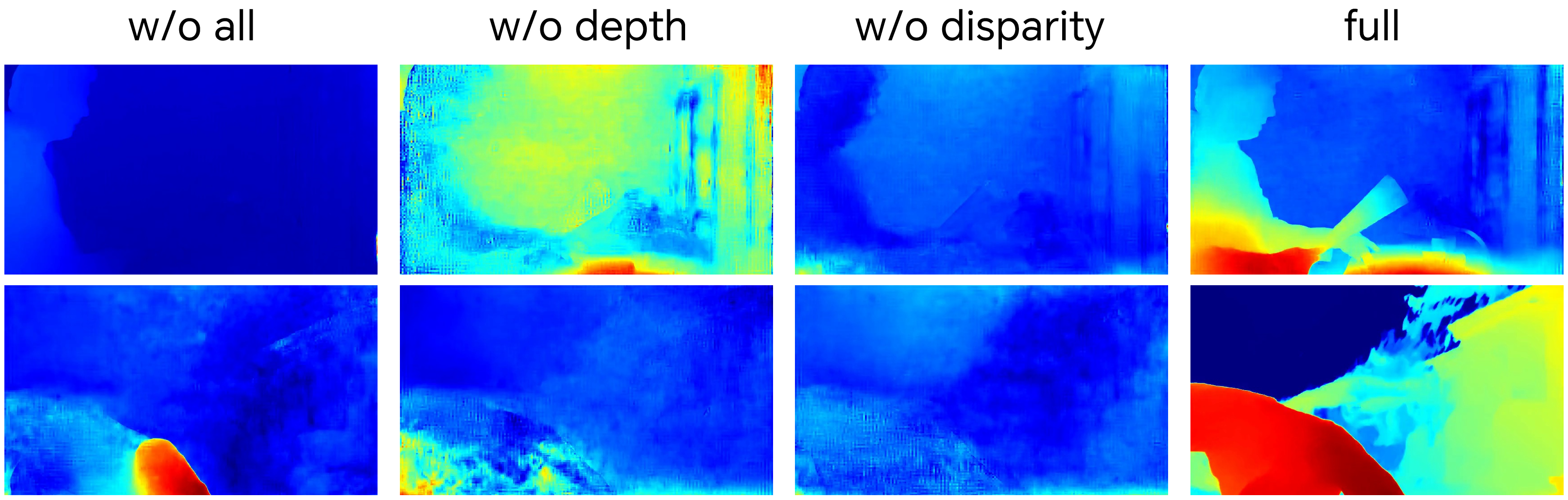} 
    \caption{\textbf{Qualitative comparison results of ablation study.} Our full model exhibits better disparity shifts and structural perception.} 
    \label{fig:ablation} 

\end{figure}

\subsection{Ablation Studies}
We analyze the contributions of geometry-aware regularization during training. As shown in Fig.~\ref{fig:ablation} and Tab.~\ref{tab:ablation}, removing either component leads to noticeable degradation in both visual fidelity and geometric accuracy. The the disparity supervision enhances disparity magnitude, while depth supervision improves depth boundary perception and spatial structure. The full model achieves the best overall performance, demonstrating that these two complementary supervision signals are essential for producing geometrically accurate and visually faithful videos.

\begin{table}[tbp]
    \centering
    \footnotesize                               
    \setlength{\tabcolsep}{2pt}
    \begin{tabular}{cc|ccccc} 
        \toprule
        Depth Sup. & Disp. Loss & PSNR $\uparrow$ & SSIM $\uparrow$ & LPIPS $\downarrow$ &
        EPE $\downarrow$ & D1-all $\downarrow$ \\
        \midrule
        \ding{55} & \ding{55}  & 23.413 & 0.742 & 0.152 & 42.318 & 0.613 \\
        \ding{51} & \ding{55}  & 24.104 & 0.758 & 0.132 &  37.593 & 0.574 \\
        \ding{55}  & \ding{51} & 24.509 & 0.781 & 0.113 &  29.998 & 0.522 \\
        \ding{51} & \ding{51} & \textbf{25.979} & \textbf{0.796} & \textbf{0.095} &  \textbf{17.453} & \textbf{0.421} \\
        \bottomrule
    \end{tabular}
    \caption{Ablation on geometry-aware regularization. The full model achieves the best overall performance.}
    \label{tab:ablation}

\end{table}

\begin{table}[tbp]
    \centering
    \begin{tabular}{l c c c c} 
        \toprule 
        Method & SE $\uparrow$ & VQ $\uparrow$ & BC $\uparrow$ & TC $\uparrow$ \\
        \midrule 
        GenStereo \cite{qiao2025genstereo} & 3.8 & 3.6 & 4.3 & 3.7 \\
        SVG \cite{dai2024svg} & 4.0 & 3.9 & 3.9 & 4.1 \\
        StereoCrafter \cite{zhao2024stereocrafter}  & 4.2 & 4.0 & 4.1 & 4.2\\
        Ours & \textbf{4.8} & \textbf{4.7} & \textbf{4.9} & \textbf{4.8}\\
        \bottomrule 
    \end{tabular}
     
    \caption{Results of Human evaluation with metrics: Stereo Effect (SE), Visual Quality (VQ), Binocular Consistency (BC), and Temporal Consistency (TC).}
    \label{tab:human}

\end{table}

\subsection{Human Evaluation}

To further assess perceptual quality, we conducted a human evaluation with 20 participants who rated 15 generated scenes. Following the protocol of SVG~\cite{dai2024svg}, participants scored each scene on four aspects using a 1–5 scale: Stereo Effect (SE), Visual Quality (VQ), Binocular Consistency (BC), and Temporal Consistency (TC).
Specifically, Stereo Effect measures the perceived 3D depth and immersion in XR displays, Visual Quality assesses image clarity and realism, Binocular Consistency evaluates alignment between left and generated right views, and Temporal Consistency reflects frame-to-frame stability over time.
As summarized in Tab.~\ref{tab:human}, our method achieves the highest scores across all subjective dimensions. Participants consistently reported that StereoWorld delivers more natural depth perception, fewer cross-view mismatches, and smoother motion continuity compared to other approaches, validating its superior perceptual and stereoscopic experience.

\section{Conclusion and Limitations}
We present StereoWorld, an end-to-end diffusion-based framework for monocular-to-stereo video generation that produces high-quality results with strong visual consistency and geometric accuracy between left and right views. The model is further optimized for long-duration and high-resolution videos, demonstrating significant practical potential.
Nevertheless, our approach has limitations. The disparity is learned in an end-to-end manner, limiting explicit control over the stereo \textit{baseline}, and the current generation speed is relatively slow, requiring around six minutes per clip. Future work will explore model distillation and other acceleration strategies to improve efficiency and expand real-world applicability.

\clearpage
{
    \small
    \bibliographystyle{ieeenat_fullname}

    \bibliography{main}
}


\end{document}